%% file: main.tex
\documentclass[letterpaper, 10pt, conference]{IEEEtran}

\usepackage{cite}
\usepackage{float}
\usepackage{amsmath}
\usepackage{amssymb}
\usepackage{amsfonts}
\usepackage{array}
\usepackage{tabularx}
\usepackage{breqn}
\usepackage{url}
\usepackage[colorlinks=true, allcolors=black]{hyperref}
\usepackage{csvsimple}
\usepackage{caption}
\usepackage{tikz}
\usepackage{tikzscale}
\usepackage{svg}
\usetikzlibrary{arrows}
\usepackage{multirow}
\usepackage{algorithm}
\usepackage{algpseudocode}
\usepackage{algcompatible}
\usepackage[english]{babel}
\usepackage{amsthm}
\newtheorem{theorem}{Theorem}
\usepackage{graphicx}
\usepackage{subcaption}
\usepackage{scalefnt}
\usepackage{capt-of}
\usepackage{booktabs}
\usepackage{comment}
\usepackage{multimedia}
\usepackage{lastpage}
\usepackage{fancyhdr}
\usepackage{siunitx}

\input{utils/functions}

\begin{document}


\title{\vspace*{0.2in} Tracking the Invisible: Self-Supervised Wheel Track Detection with Thermal-RGB Fusion}

\author{
    Yiming~Yang$^*$,
    Jeremy~P. Bos$^\ddagger$
    \thanks{$^*$$^\dagger$$^\ddagger$ Y. Yang and J. Bos are with the Department of Electrical and Computer Engineering, Michigan Technological University, Houghton,
    MI, USA.}
    \thanks{$^*$yimingya@mtu.edu, $^\ddagger$jpbos@mtu.edu}
}

\maketitle

\thispagestyle{empty}

\begin{abstract}

Accurate detection of drivable area is essential for autonomous driving and advanced driver assistance systems. While state-of-the-art methods focus on general traversability estimation, they prove inadequate for snow-covered roads where following existing wheel tracks offers crucial advantages: improved traction, predictable vehicle dynamics, and verified path stability. To address this, we propose a low-cost, self-supervised approach detecting safety-critical wheel tracks using synchronized RGB-I (RGB + Long-Wave Infrared) imagery. Our method automatically generates training labels by projecting future wheel trajectories from recorded GNSS data. A neural network is designed to identify wheel tracks with the 4-channel RGB-I images from challenging, white-out road surfaces. Experimental results demonstrate the effectiveness of our approach in visually monotonous and featureless adverse environment, offering a robust and cost-effective alternative for winter autonomy. The newly introduced ThermalTrack dataset is available at: \url{https://digitalcommons.mtu.edu/thermaltrack/}.

\end{abstract}


\IEEEpeerreviewmaketitle



\section{Introduction}
\label{sec:introduction}
\input{sections/introduction.tex}

\section{Related Works}
\label{sec:relatedworks}
\input{sections/relatedworks.tex}

\section{Methods}
\label{sec:methods}
\input{sections/methods.tex} 

\section{Experiments}
\label{sec:experiments}
\input{sections/experiments.tex}

\section{Conclusion}
\label{sec:conclusion}
\input{sections/conclusion.tex} 

\def\IEEEbibitemsep{0pt plus .5pt}
\bibliographystyle{IEEEtran}
\bibliography{bibliography/references}

\ifCLASSOPTIONcaptionsoff
  \newpage
\fi

\end{document}

%% file: utils/functions.tex
\newtheorem{lemma}[theorem]{Lemma}

\newcolumntype{C}{>{\centering\arraybackslash}X}



%% file: sections/introduction.tex
\begin{figure}[htp]
    \centering
    \includegraphics[width=8.2cm]{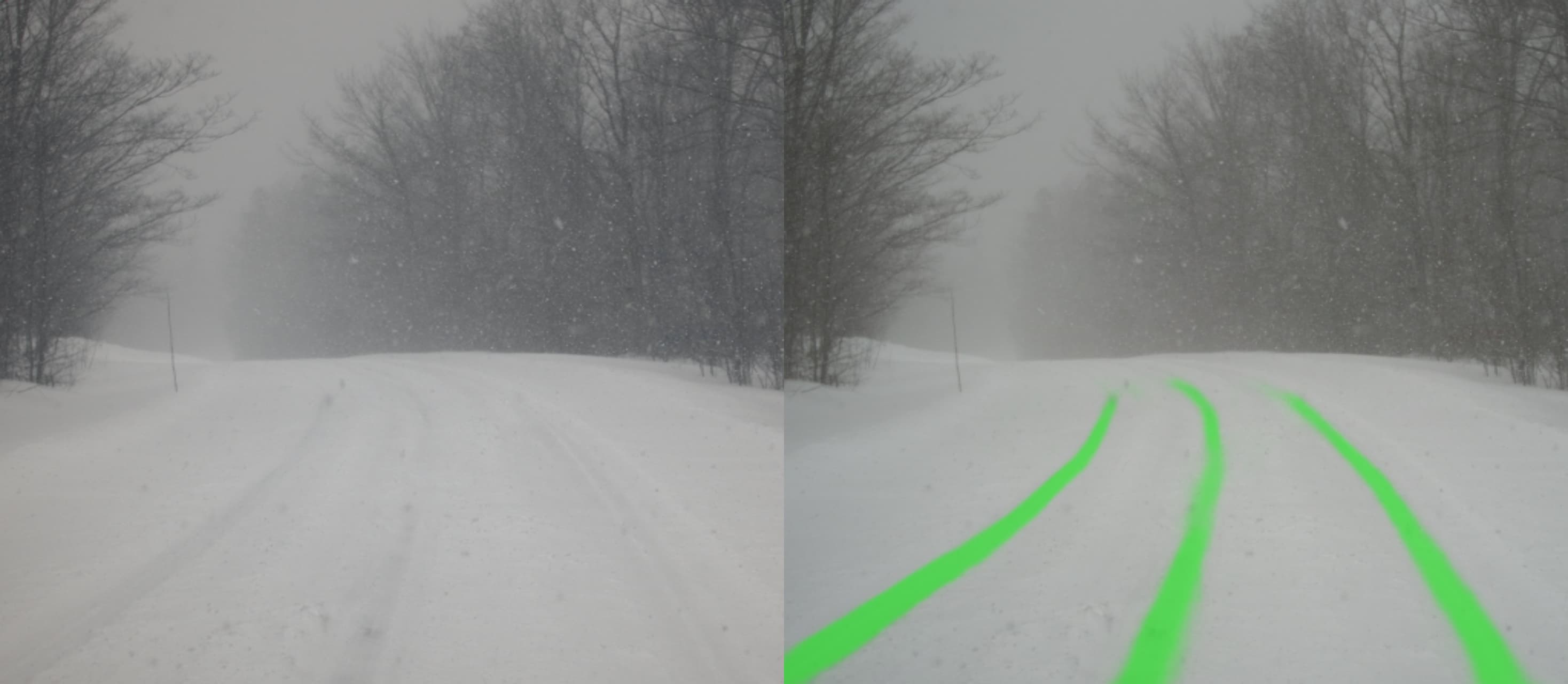}
    \caption{Left: example from WADS showing severe snow occlusion: road surfaces, lane markings, and structural boundaries are nearly indistinguishable, with only subtle terrain variations hinting at the underlying roadway. Right: the predicted probability map for wheel tracks.}
    \label{fig:wheel_track_example}
    \vspace{-5mm}
\end{figure}

\IEEEPARstart{A}{utonomous} driving in cold regions faces significant challenges on snow-covered roads, where lane lines and road structures are often obscured by snow. For traditional navigation systems, Ruotsalainen et al.\cite{9115644} show that existing fusion and map-matching algorithms for vehicle localization are inadequate for navigation in Arctic environments. LiDAR and HD map-based perception systems also experience localization loss under snow-covered road conditions\cite{USDOT_AVAW2021}. Furthermore, HD maps struggle in these conditions due to altered driving dynamics: vehicles carving new paths in the snow shift lane centers and modifying lane widths beyond their original boundaries, making pre-mapped data unreliable. Similarly, the dynamic nature of adverse environments, where features change or disappear, disrupts map consistency and localization. Vision-based navigation struggles also in colorless and featureless environments due to challenges in visual feature extraction, leading to failures in visual odometry and navigation\cite{8931615}. Fig. \ref{fig:wheel_track_example} shows a white-out frame from Winter Adverse Driving dataSet (WADS)\cite{kurup2023winter}, with the corresponding wheel-track probability map predicted by our method. LiDAR SLAM encounters difficulties in distinguishing between snow-covered surfaces and actual road features, leading to unstable localization and navigation errors\cite{Baril_2022}. These challenges highlight the need for robust perception systems capable of adapting to dynamic and unstructured environments.

In such unstructured environments, wheel tracks carved into the snow offer a viable alternative for navigation. Unlike obscured lane markings or transient visual features, wheel tracks provide physical guidance that persists even in heavy snow. Research has shown that aligning a vehicle's wheels with existing ruts during off-road driving offers multiple advantages\cite{ordonez2010rut}. First, it enhances safety by reducing lateral slippage during turns, as the ruts provide extra lateral force to guide the vehicle along its path. It also prevents the ruts from acting as obstacles, which reduces undesirable vibrations and instability. It increases traction, particularly when traversing loose terrains such as mud, sand, and snow. This strategy also saves energy by reducing the energy wasted on compacting the ground.

Traversability estimation\cite{8594031}, a critical component of indoor/outdoor navigation, has gained considerable attention and made significant progress, both in urban\cite{9561394} and rural\cite{9839522} environments. These methods typically focus on identifying drivable areas by classifying terrain as traversable or non-traversable, leveraging advances in machine learning, sensor fusion, and traveling experience. While they excel at detecting general drivable regions, they fall short in snow-covered driving scenarios: they fail to account for the unique driving behavior in snow, where drivers tend to follow existing wheel tracks for safer and more stable travel. This gap highlights the need for a more specialized approach that focuses on wheel track detection over generic drivable area estimation, ensuring more robust navigation in adverse conditions.

In this work, we propose a novel method for detecting wheel tracks on snow-covered roads by fusing RGB and thermal images. While wheel tracks are often indistinct in RGB images, thermal imaging reveals critical temperature differentials between compacted wheel tracks (higher thermal inertia, emitting more radiation) and loose snow (higher reflectivity, emitting less radiation). To overcome the lack of clear visual features, we introduce a RGB-thermal fusion based self-supervised learning framework that mitigates manual labeling by projecting recorded wheel track trajectories onto image frames.

We validate our approach using the WADS dataset, with experiments demonstrating robust wheel track detection under extreme visual degradation (snow occlusion, low contrast, and homogeneous textures). Our method shows consistent improvements across standard evaluation metrics compared to baseline approaches. This work advances winter autonomy by leveraging thermal sensing and integrating wheel track detection with state-of-the-art traversability estimation technology.

%% file: sections/relatedworks.tex
\subsection{Self-supervised Traversability Estimation}

Traversability estimation plays a vital role in enabling autonomous navigation across diverse environments. Early approaches relied on analyzing basic features like slope, roughness, elevation\cite{7759199}, and vision\cite{7487782}. The rise of deep learning has shifted focus toward semantic segmentation techniques, such as terrain classification\cite{shaban2022semantic} and image segmentation\cite{Zhao_2017_CVPR}. Despite their effectiveness, these methods depend heavily on annotated datasets, limiting their ability to adapt to unfamiliar or dynamic conditions\cite{barnes2017find}. This has driven the development of self-supervised techniques, which learn from real-world interactions without extensive labeling, offering greater robustness and adaptability in uncertain environments.

 In paper\cite{10146445}, Seo et al. introduced a one-class classification method to distinguish traversable from non-traversable areas in visually diverse environments. Other approaches, such as employing autoencoders\cite{schmid2022self} to identify low- and high-risk terrains based on reconstruction errors, or using normalizing flow\cite{8963641} for traversability classification on multi-modal images, have also demonstrated strong performance.

\subsection{Wheel Tracks Detection}

LiDAR and cameras are widely used for wheel track detection, typically relying on geometry-based feature extraction. Ordonez \cite{ordonez2011laser} highlighted the advantages of rut tracking for off-road autonomous vehicles, designing a laser range finder based system to detect and follow ruts. Zhang \cite{zhang2024identification} created a hand-crafted image-processing pipeline for identifying wheel tracks in wheat fields, incorporating feature filtering and extraction, curve and polynomial fitting. Fuentes et al. \cite{fuentes2021quantification} introduced an automated LiDAR method for rut detection using digital elevation models. However, depending on manually made ruts could reduce its performance in actual challenging environments Seok et al. \cite{seok2024lidar} proposed a LiDAR-based bump and dip detection method for paved road, which shows potential for wheel track detection but involves multiple processing stages and highly customized filtering and feature extraction process, limiting its generalizability across diverse winter conditions. For snow-covered environments, Goberville et al. \cite{goberville2022tire} developed a supervised learning-based tire track detection system, using a custom-labeled dataset and convolutional neural network (CNN) architecture. This approach requires intensive manual annotation and lacks generalization. Kim et al.\cite{s24237802} proposed a YOLOv5-based system for snowy lane-keeping, using manually annotated tire tracks and geometric post-processing (BEV/Hough) for path generation, but it relies on labor-intensive annotations and produces suboptimal bounding boxes that requiring future refinement. While these methods demonstrate progress, they rely on traditional feature extraction and manual annotation, limiting their adaptability to dynamic and complex environments. This highlights the need to integrate cutting-edge technology from adjacent research fields to achieve robust, generalizable performance.

\subsection{IR Thermal Camera based Navigation}

IR thermal imaging provides a low-cost, safe, and effective solution for mobile robot perception. Xaud et al.\cite{xaud2019thermal} addressed sugarcane robot navigation by combining thermal imaging with CNN-based segmentation to identify crop rows and compute centerline offset, outperforming RGB methods and offering a cost-effective alternative to Lidar-based systems. To tackle nighttime autonomous navigation challenges, Aditya et al.\cite{aditya2024thermal} developed a system combining thermal cameras with model predictive control. By fusing thermal and visible modalities during training, their approach achieves performance comparable to daytime RGB-based systems while operating in low-light conditions. For research that leveraging thermal image in adverse weather conditions, Vachmanus et al.\cite{vachmanus2021multi} introduced an RGB-thermal fusion framework for robust road and pedestrian segmentation in snow, achieving better performance than existing approaches via a dedicated fusion module and pyramid supervision. Renaux\cite{9294671} et al. improved autonomous vehicle perception in harsh weather by integrating thermal imaging with RGB data. Their method fused thermal intensity analysis with YOLOv3-tiny detections for reliable person detection, while a modified MultiNet achieved accurate road segmentation with minimal training.

\subsection{Research Gap}

Existing research on wheel track detection has primarily focused on scenarios where wheel tracks exhibit relatively consistent and well-defined characteristics, such as those found in wheat fields or manually created tracks. However, detecting wheel tracks on snow-covered surfaces introduces unique challenges due to the highly variable and complex nature of the tracks. In snowy environments, wheel tracks often overlap, vary in depth, and exhibit irregular shapes due to factors such as human inaccuracy in following existing tracks, new snowfall, and wind effects. These problems, which are not typically encountered in more controlled environments demand advanced techniques to accurately identify and distinguish wheel tracks in such dynamic and visually ambiguous conditions.

%% file: sections/methods.tex
We present a low-cost neural network based solution that addresses the critical challenge of winter navigation: reliably identifying wheel tracks in visually degenerate snow environments. Our approach exploits subtle thermal and textural signatures in compacted snow to guide autonomous systems through otherwise featureless terrain.

\subsection{Wheel Tracks Labeling}
\begin{figure}[htp]
    \centering
    \includegraphics[width=8.2cm]{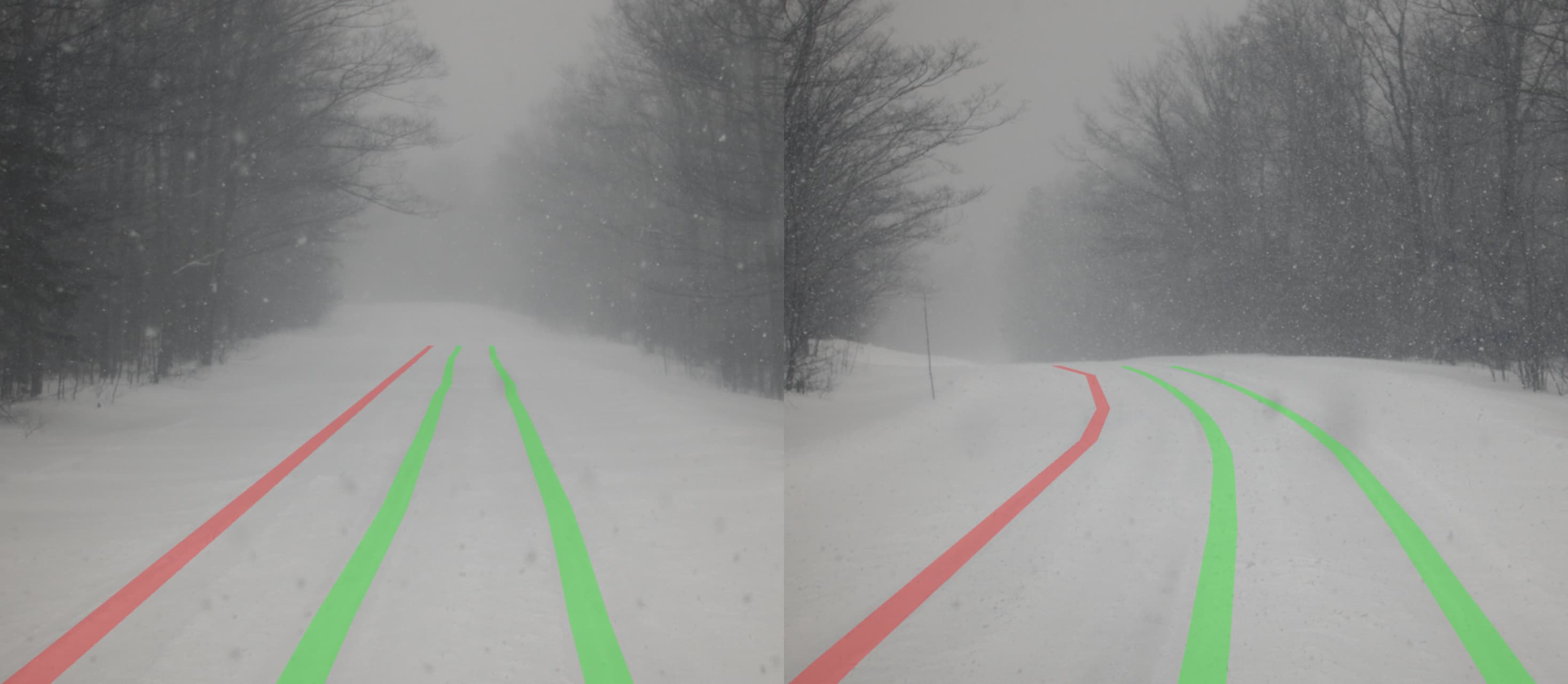}
    \caption{Self-supervised wheel track labels generated by our automated procedure. One challenge in wheel track detection is that tracks appear not only in the ego vehicle’s travel path but also in the opposite path. Ignoring these can degrade training performance. The images show projected wheel track trajectories: (1) self-supervised tracks computed from GNSS data using vehicle geometry and transformations (green), and (2) manually labeled tracks from the opposite travel path (red).}
    \label{fig:wheel_track_labeling}
    \vspace{-3mm}
\end{figure}

Similar to the research works\cite{10146445,schmid2022self,alamikkotervo2024tadap}, the wheel track trajectories are projected onto images as training labels by an automated procedure, as illustrated in Fig. \ref{fig:wheel_track_labeling}. In our setup, the wheel tracks in the environment were left by prior traffic before data collection. The ground truth labels are derived by projecting the ego-vehicle's own GNSS measurements onto the images as it drives along these pre-existing tracks during the compilation of the WADS dataset. The horizon-$h$ trajectories starting at time $t$ are converted into wheel track contact points, denoted as $P(t, t+h)$. These contact points are then projected into camera coordinates using the following equations to generate self-supervised labels $l$, as this labeling process inherently relies on synchronized, cross-modal sensor data mapping the GNSS trajectory directly to the camera frame.

\begin{equation}
    l = K\cdot[R|T]\cdot{P(t, t+h)}
\end{equation}

Here, $K$ denotes the intrinsic camera calibration matrix, and $[R\mid T]$ represents the extrinsic transformation matrix. $R$ denotes the rotation matrix and $T$ represents the translation vector. Unlike\cite{10146445,schmid2022self}, which generates dense positive labels by marking all pixels between left and right wheel-terrain contact points while leaving other regions unlabeled, our method exclusively labels the precise wheel-terrain contact points as positive. This results in an extremely sparse positive label set, accounting for less than 5\% of pixels on average, creating a significant class imbalance that poses challenges for model training.

Beyond label sparsity, wheel track detection introduces unique complexity compared to normal drivable area segmentation: tracks appear not only in the ego vehicle's path but also in opposite traffic lanes. If not handled properly, this can introduce ambiguity during training and affect model performance. For wheel tracks in the ego vehicle’s path, we use our automated labeling method. However, since no corresponding trajectory labels are available for wheel tracks left by oncoming vehicles, we supplement the dataset with manually annotated labels using a labeling tool. This hybrid approach ensures that the model learns to detect all relevant wheel tracks while maintaining the benefits of self-supervision for ego-vehicle tracks. On average, the manual annotation of oncoming wheel tracks required approximately 8 to 10 seconds per image. For the entire dataset of 951 images before augmentation, the total manual post-processing time was roughly 2.5 hours.

\subsection{Input Modalities and Registration}

While utilizing a native panchromatic sensor can mitigate Bayer pattern interpolation and maximize spatial resolution, our framework is designed around the widely adopted WADS dataset, which natively provides RGB alongside LWIR data. We deliberately preserve all three RGB channels to form a four-channel ($RGB\text{-}I$) input tensor for two primary reasons. First, the multi-spectral variations across the R, G, and B channels retain critical surface texture and chromatic features. During winter driving, wheel tracks frequently transition between compacted snow, slush, ice-water mixtures, and exposed asphalt; preserving RGB data is vital for capturing these subtle material boundaries, which a panchromatic channel would compress into indistinct grayscale values. Second, retaining the standard RGB spectrum ensures seamless compatibility with state-of-the-art perception backbones, leveraging robust, large-scale feature representations that are unavailable for single-channel monochrome inputs.

Our framework processes a fused multi-spectral input consisting of RGB and LWIR thermal data. A key challenge in this fusion is spatial registration due to differing optical centers and resolutions. As illustrated in Fig. \ref{fig:system_diagram}, we downsample both modalities to $512\times512$ resolution for computational efficiency, and leveraging the camera rig's rigid, forward-facing geometry to apply a static homography transformation for alignment. While static homography can theoretically introduce minor parallax artifacts for vertically prominent 3D structures, its impact here is negligible; our primary region of interest, the driving path and road surface, closely approximates a flat, 2D plane relative to the camera baseline. This assumption yields stable alignment and sufficient fusion quality for surface-level tracking without requiring computationally expensive dynamic depth estimation or online registration frameworks.

\subsection{Wheel Tracks Learning}

\begin{figure*}[t] 
  \centering
  \includegraphics[width=0.95\textwidth]{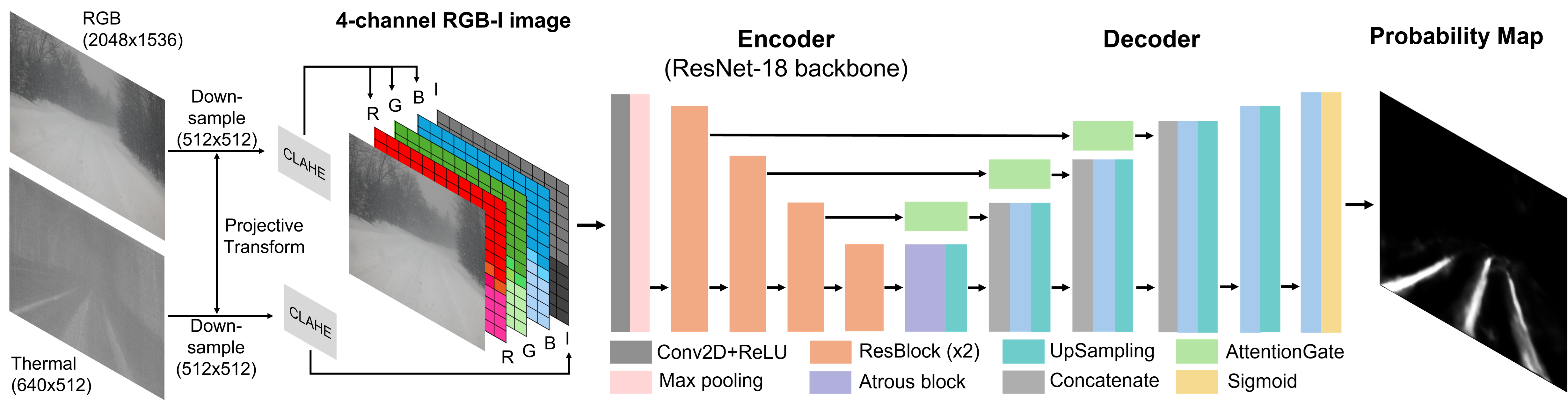} 
  \caption{The neural network architecture processes input images through the following sequential stages: (a) The RGB image and thermal image are downsampled to 512 x 512 pixels. Since the RGB-LWIR cameras rig has fixed geometry, we then employ a static homography transformation for image alignment. Contrast Limited Adaptive Histogram Equalization (CLAHE) is applied to enhance the contrast of aligned RGB and thermal images. (b) The intensity channel from the LWIR thermal image is integrated as the 4th channel into the RGB image, forming a 4-channel RGB-I input. (c) Our network architecture employs a ResNet-18 backbone encoder with Attention Gates and atrous (dilated) convolutional blocks. This design enables selective focus on wheel track regions while actively suppressing irrelevant areas. (d) The network outputs a pixel-wise probability map, where each value $\in$ [0,1] represents the predicted likelihood of the corresponding image pixel belonging to a wheel track.}
  \label{fig:system_diagram}
  \vspace{-3mm}
\end{figure*}

The overall architecture of our system is shown in Fig. \ref{fig:system_diagram}. We use an encoder-decoder structure. A combined 4 channel RGB-I image is forwarded to a feature encoder for feature extraction and contextual understanding. The backbone encoder is ResNet\cite{he2016deep}, which extracts hierarchical features through its deep residual layers, enabling multi-scale receptive fields for context aggregation. As noted in Section A, the severe class imbalance (as visually demonstrated in Fig. \ref{fig:wheel_track_labeling}) between wheel track pixels and background regions presents significant challenges for model training.

To address the extreme class imbalance in our ResNet backbone, we integrate two key components: (1) attention gates\cite{oktay2018attention} that dynamically highlights wheel-track regions in RGB-I imagery, and (2) multi-scale dilated convolutions\cite{chen2018encoder} for contextual feature extraction.

\begin{equation}
\begin{aligned}
g &= \sigma(\psi(\text{ReLU}(W_g * D + W_e * E))) \\
\tilde{E} &= g \odot E
\end{aligned}
\end{equation}

where $g$ is a spatial attention map 
that adaptively weights encoder features ($E$) to amplify wheel-track pixels and suppress background. $D$ represent decoder features, $W_g$ and $W_e$ are 1×1 convolutional kernels, $\psi$ generates spatial attention weights via sigmoid activation ($\sigma$), and $\odot$ denotes element-wise multiplication. The output $\tilde{E}$ is attention-weighted encoder features to recover spatial details while maintaining attention-guided focus. This mechanism performs channel-spatial weighting to amplify relevant wheel-track features. For multi-scale context, we employ parallel atrous convolutions:

\begin{equation}
F_{\text{ms}} = \Big\Vert_{i=1}^3 \left( W_i \ast_{r_i} X \right)
\end{equation}

where $F_{ms}$ is the multi-scale feature representation, $*_{r_i}$ denotes dilated convolution with rates $[r_{1},r_{2},r_{3}]$, and $||$ indicates channel-wise concatenation to retains all scale-specific features. This preserves fine details while capturing wheel-tracks at varying scales through expanded receptive fields. The combined approach efficiently focuses learning on sparse positive pixels while suppressing background interference.

\subsection{Loss Functions}

 While the attention gate and atrous block architecture extracts key features, we further apply focal loss to mitigate class imbalance and boundary loss to enforce geometric precision. The attention gate and focal loss jointly address label sparsity, while atrous convolutions and boundary loss ensure accurate localization across different scales.

Focal loss\cite{lin2017focal} is used to address severe class imbalance by focusing on hard-to-classify pixels and down-weighting the loss contribution from easy-to-classify pixels. The loss function is defined as:

\begin{equation}
    l_{f} = -(1 - p_t)^\gamma \log(p_t),
\end{equation}

where $\gamma$ is the focusing parameter that reduces the loss contribution from easy-to-classify pixels. $p_t$ is the model's estimated probability for the true class:
    \[
    p_t = 
    \begin{cases} 
    p & \text{if } y = 1 \\
    1 - p & \text{if } y = 0,
    \end{cases}
    \]

Boundary loss\cite{kervadec2019boundary} ensures accurate segmentation along the edges of wheel tracks by penalizing discrepancies between the predicted and projected wheel tracks' boundaries. The loss function is defined as:

\begin{equation}
    l_{b} = \frac{1}{N} \sum_{i=1}^N \|\nabla y_{\text{true}}^{(i)} - \nabla y_{\text{pred}}^{(i)}\|^2,
\end{equation}

where $\nabla y_{\text{true}}$ and $\nabla y_{\text{pred}}$ are the gradients of the ground truth and predicted segmentation maps, respectively, $N$ is the total number of pixels. The total loss function combines focal loss and boundary loss to address the class imbalance and improve boundary accuracy simultaneously. It is defined as:

\begin{equation}
    l = w_1{l_{f}} + w_2{l_{b}},
\end{equation}

where $w_1$ and $w_2$ are the weights for focal loss and boundary loss, respectively. Through empirical evaluation, we select $w_1 = 1$ and $ w_2 = 1$ to balance the contributions of both losses.

%% file: sections/experiments.tex
In this section, we validate our self-supervised wheel track detection method under diverse, challenging winter conditions, including white-out scenarios with severely degraded visibility, snow surfaces lacking visual contrast, and variable track geometries. We first describe the dataset we used for this work, then followed by the experimental setup and evaluation metrics. In the experiment results, we provide both quantitative and qualitative analysis. Finally, we present ablation studies demonstrating that: (1) thermal imaging plays a critical role in wheel track detection by enhancing features in low-contrast scenarios where RGB-only method fails, and (2) the integration of attention gates and atrous convolutional blocks yields significant performance gains by jointly optimizing multi-scale context aggregation and spatially adaptive feature refinement.

\subsection{Dataset}
We utilize WADS, a specialized dataset designed for autonomous vehicles and robot research in inclement winter weather. WADS focuses exclusively on challenging conditions such as heavy snowfall and occasional white-out scenarios, providing a robust platform for testing and development. The dataset includes data collected from high-resolution LiDAR, visible and infrared cameras, and Global Navigation Satellite Systems (GNSS) coupled with Inertial Measurement Units (IMUs). Notably, WADS is one of the few datasets that emphasizes rural and semi-rural environments, making it uniquely suited for studying winter driving in less urbanized settings. The WADS dataset comprises images captured across diverse conditions, encompassing variations in road surfaces, weather, lighting, and terrain types, etc. 

From this raw multi-modal data, we curate a specialized sequence focused explicitly on winter path traversability, which we detail in Section III-A and introduce as the ThermalTrack dataset. To enhance the robustness and generalization of our model, we performed dataset augmentation using techniques including random flip, rotation, brightness, contrast, scaling, noise injection, and perspective change to simulate a wider variety of challenging road conditions and improve the model's ability to handle real-world variability. We augmented 951 images to 7608 in total for model training.

\subsection{Experimental Setup}
All the models are trained for 10 epochs with a batch size of 4 and an image size of 512 x 512, using the Adam optimizer with a learning rate scheduler starting at 1e-4, as well as an L2 regularizer. To establish a rigorous evaluation benchmark, we select 4 established neural network architectures that have been extensively validated in research for image segmentation and related structured prediction tasks as baselines: (1) UNet: The predominant architecture for biomedical image segmentation\cite{ronneberger2015u}, adopted here for its demonstrated efficacy in fine-grained localization tasks. (2) PSPNet: Originally developed for scene parsing\cite{zhao2017pyramid}, selected for its systematic approach to multi-scale context aggregation through pyramid pooling. (3) DeepLabv3+: The prevailing architecture in recent autonomous driving benchmarks\cite{chen2018encoder}, notable for its hybrid approach combining atrous convolution and encoder-decoder refinement. (4) ResNet18: Implemented in both standard form and our modified variant (Ours) to quantify the contribution of our architectural innovations.

To verify the effectiveness of attention gate and atrous blocks, our architectural modifications were carefully designed to complement each baseline's inherent capabilities: For UNet and ResNet18, we integrated both attention gates and atrous blocks to enhance feature selectivity and multi-scale context aggregation, as these architectures lack native pyramid representations. For PSPNet, we preserved its native pyramid pooling module (PPM) while adding attention gates to refine feature weighting, avoiding redundant multi-scale processing that could degrade its hierarchical feature fusion. For DeepLabv3+, we maintain its atrous spatial pyramid pooling (ASPP) but introduce attention gates to optimize feature selection from parallel dilation branches, preventing information overload in its existing multi-scale pipeline.

\subsection{Evaluation Metrics}
To quantitatively assess the performance of the methods, We adopt the evaluation framework from \cite{10146445}, employing two complementary metric classes: Area Under the Receiver Operating Characteristic (AUROC) and standard evaluation metrics from the KITTI road detection benchmark. The AUROC measures the probability that a positive instance will receive a higher normality score than a negative instance, providing a threshold-independent evaluation of one-class classification algorithms. The standard evaluation metrics from the KITTI road detection benchmark include: Maximum F1-score (MaxF), Average Precision (AP), Precision (PRE), Recall (REC), False Positive Rate (FPR), False Negative Rate (FNR). Note that PRE, REC, FPR, and FNR are computed at the threshold corresponding to the maximum F1-score.

\subsection{Experiment Results}
\begin{figure*}[t] 
  \centering
  \includegraphics[width=0.92\textwidth]{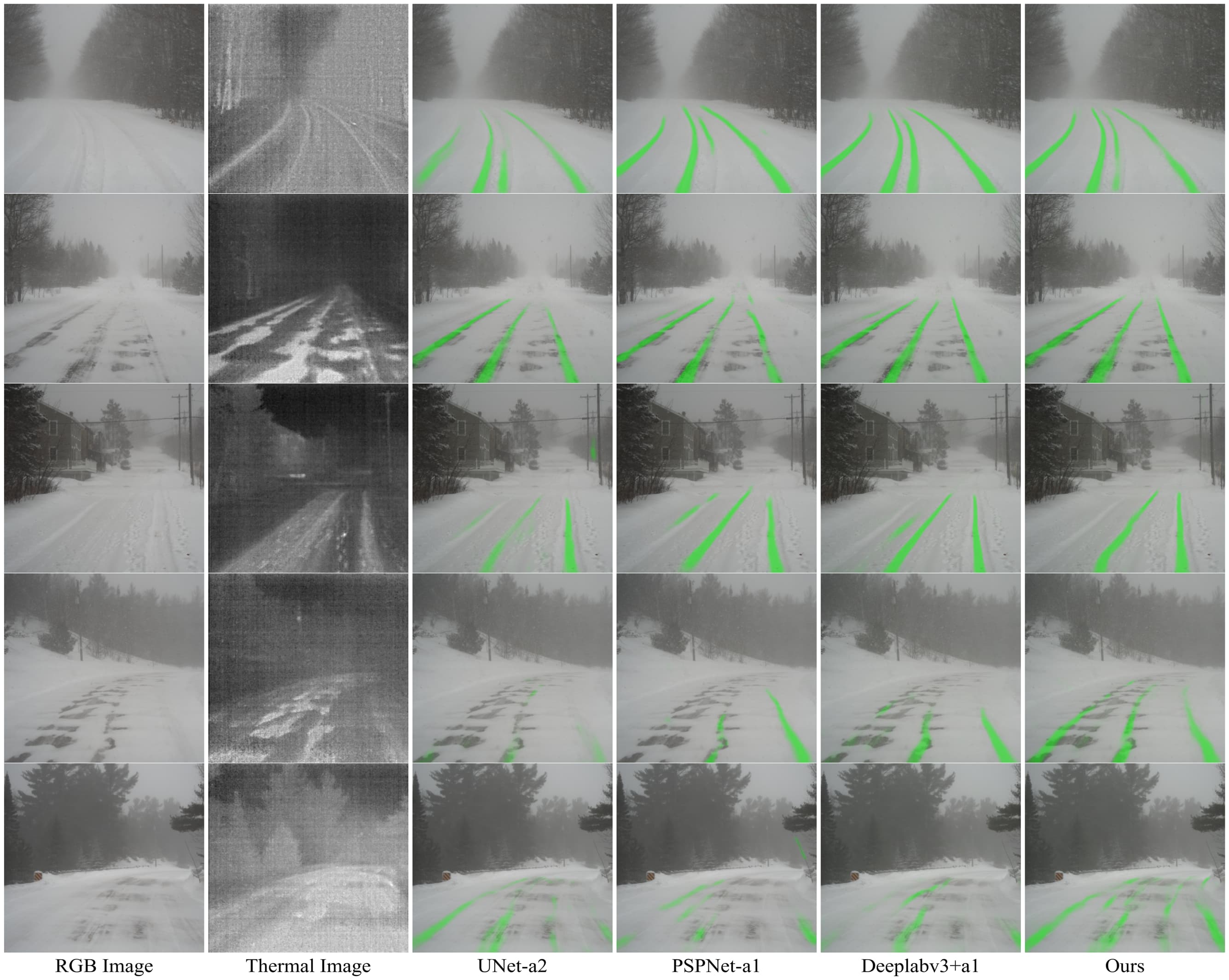} 
  \caption{Qualitative wheel track detection results and probability maps visualize model performance. As illustrated in the final row, the proposed method exhibits limitations in certain scenarios: in cases where wheel track features are absent in both RGB and thermal imagery, the network produces fragmented and low-confidence predictions.}
  \label{fig:qualitative_results_models}
  \vspace{-3mm}
\end{figure*}

Quantitative and qualitative results for our dataset are presented in Table \ref{tab:performance_comparison_networks} and \ref{tab:performance_comparison_inputs}, and Fig. \ref{fig:qualitative_results_models}, respectively. All the models are trained in a self-supervised manner with RGB-I images except for Ours-rgb and Ours-i. Our method achieves superior performance across most categories compared to existing approaches, indicating its effectiveness in generating more confident and reliable predictions.

\begin{table}[t]
	\centering
	\caption{Quantitative results of wheel track detection using baseline models, all the models are trained on RGB-I images.}
	\label{tab:performance_comparison_networks} 
	\setlength{\tabcolsep}{2pt} 
	\fontsize{8.5pt}{10.5pt}\selectfont
	\begin{tabularx}{\columnwidth}{Xccccccc}
		\toprule[1.0pt]
		\textbf{Model} & \textbf{AUROC} & \textbf{MaxF} & \textbf{AP} & \textbf{PRE} & \textbf{REC} & \textbf{FPR} & \textbf{FNR} \\
		\midrule
		UNet & 0.7068 & 0.5197 & 0.6079 & 0.7691 & 0.4192 & 0.0056 & 0.5808 \\
		PSPNet & 0.7532 & 0.6102 & 0.6666 & 0.7978 & 0.5122 & 0.0057 & 0.4878 \\
		DeepLabv3+ & 0.7446 & 0.6002 & 0.6629 & \textbf{0.8074} & 0.4943 & \textbf{0.0052} & 0.5057 \\
		ResNet18 & \textbf{0.8462} & \textbf{0.7202} & \textbf{0.7337} & 0.7499 & \textbf{0.7034} & 0.0109 & \textbf{0.2966} \\
		\bottomrule[1.0pt]
	\end{tabularx}
	\vspace{-3mm}
\end{table}

\begin{table}[t]
	\centering
	\caption{Ablation study for wheel track detection comparing RGB-only, Thermal-only, and RGB-I inputs, with model variants a1 (attention gate) and a2 (attention gate+atrous blocks). Ours model is ResNet18+a2. All the models are trained with RGB-I images (except Ours-rgb/Ours-i).}
	\label{tab:performance_comparison_inputs}
	\setlength{\tabcolsep}{2pt} 
	\fontsize{8.5pt}{10.5pt}\selectfont 
	\begin{tabularx}{\columnwidth}{Xccccccc}
		\toprule[1.0pt]
		\textbf{Model} & \textbf{AUROC} & \textbf{MaxF} & \textbf{AP} & \textbf{PRE} & \textbf{REC} & \textbf{FPR} & \textbf{FNR} \\
		\midrule
		UNet+a2 & 0.7843 & 0.6529 & 0.6941 & \textbf{0.7929} & 0.5753 & 0.0068 & 0.4247 \\
		PSPNet+a1 & 0.8476 & 0.7186 & 0.7337 & 0.7471 & 0.7064 & 0.0112 & 0.2936 \\
		DeepLabv3+a1 & 0.8589 & 0.7460 & 0.7584 & 0.7757 & 0.7281 & 0.0102 & 0.2719 \\
		\midrule
		Ours-rgb & 0.6025 & 0.3083 & 0.5164 & 0.6901 & 0.2073 & \textbf{0.0024} & 0.7927 \\
		Ours-i & 0.7687 & 0.5803 & 0.6127 & 0.6525 & 0.5510 & 0.0137 & 0.4490 \\
		Ours-rgbi & \textbf{0.8744} & \textbf{0.7579} & \textbf{0.7675} & 0.7633 & \textbf{0.7601} & 0.0114 & \textbf{0.2399} \\
		\bottomrule[1.0pt]
	\end{tabularx}
	\vspace{-3mm}
\end{table}

\begin{figure}[htp]
    \centering
    \includegraphics[width=8.8cm]{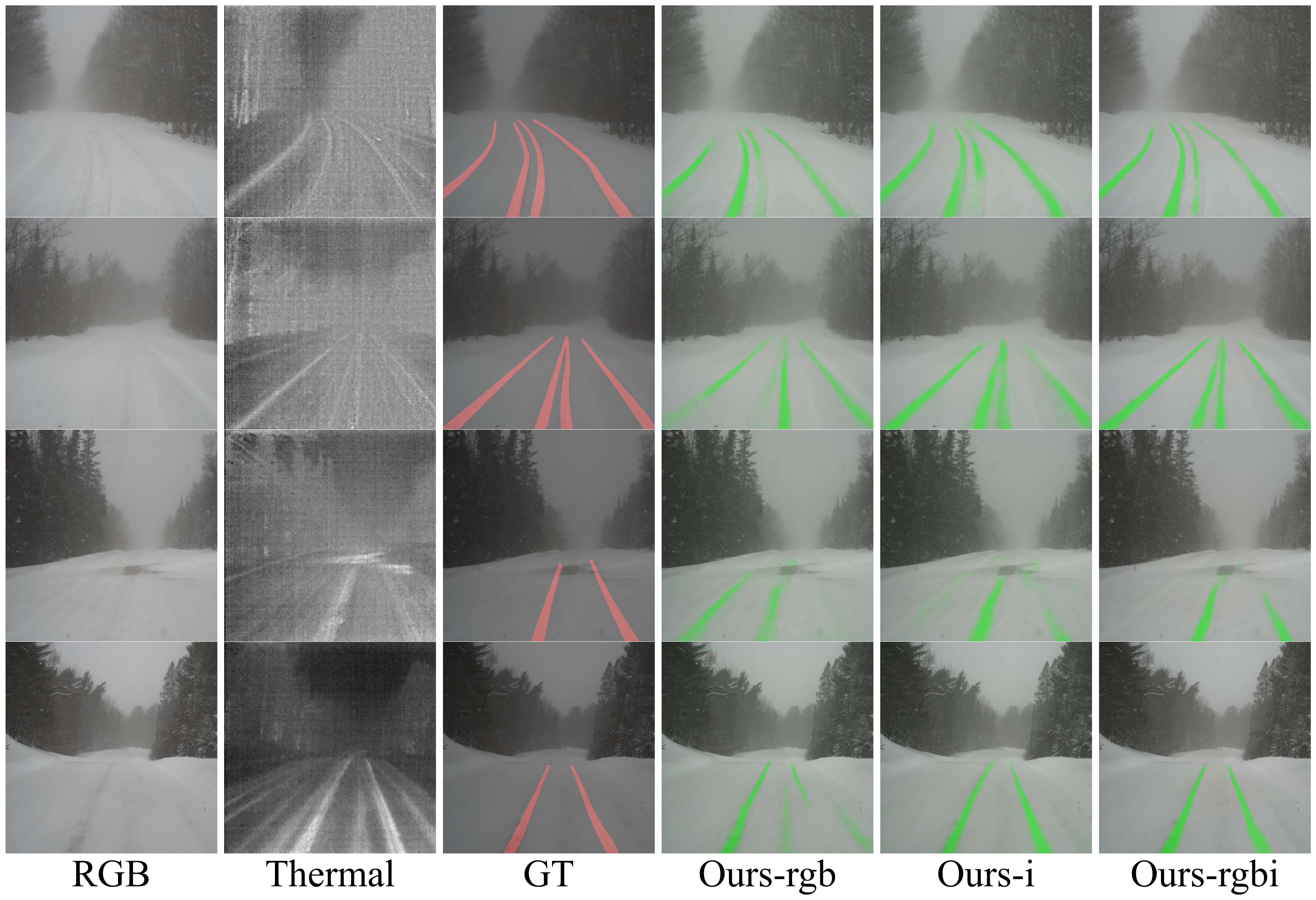}
    \caption{Qualitative results demonstrating the effectiveness of thermal imaging on wheel track detection. The models were trained independently on RGB, thermal, and RGB-thermal (RGB-I) images.}
    \label{fig:qualitative_results_inputs}
    \vspace{-3mm}
\end{figure}

For the ablation studies, we analyze the effectiveness of (1) thermal image and (2) attention gate and atrous blocks. For thermal images, from Table II, quantitative analyses reveal that thermal imagery reduces the false negative rate (FNR) by 69.74\% compared to the RGB-only baseline while maintaining a low false positive rate (0.014). The qualitative results are shown in Fig. \ref{fig:qualitative_results_inputs}, in the first case, RGB and Thermal predictions both have missing/discontinuous parts, and their fusion proves effective. In the second, Thermal finds tracks missed by RGB. The third case reveals that RGB primarily detects the more prominent wheel tracks from the opposite lane, while Thermal correctly identifies those in the correct lane. RGB produces unrealistic breaks/offsets in the fourth case, whereas Thermal remains precise.

\begin{figure}[htp]
    \centering
    \includegraphics[width=8.5cm]{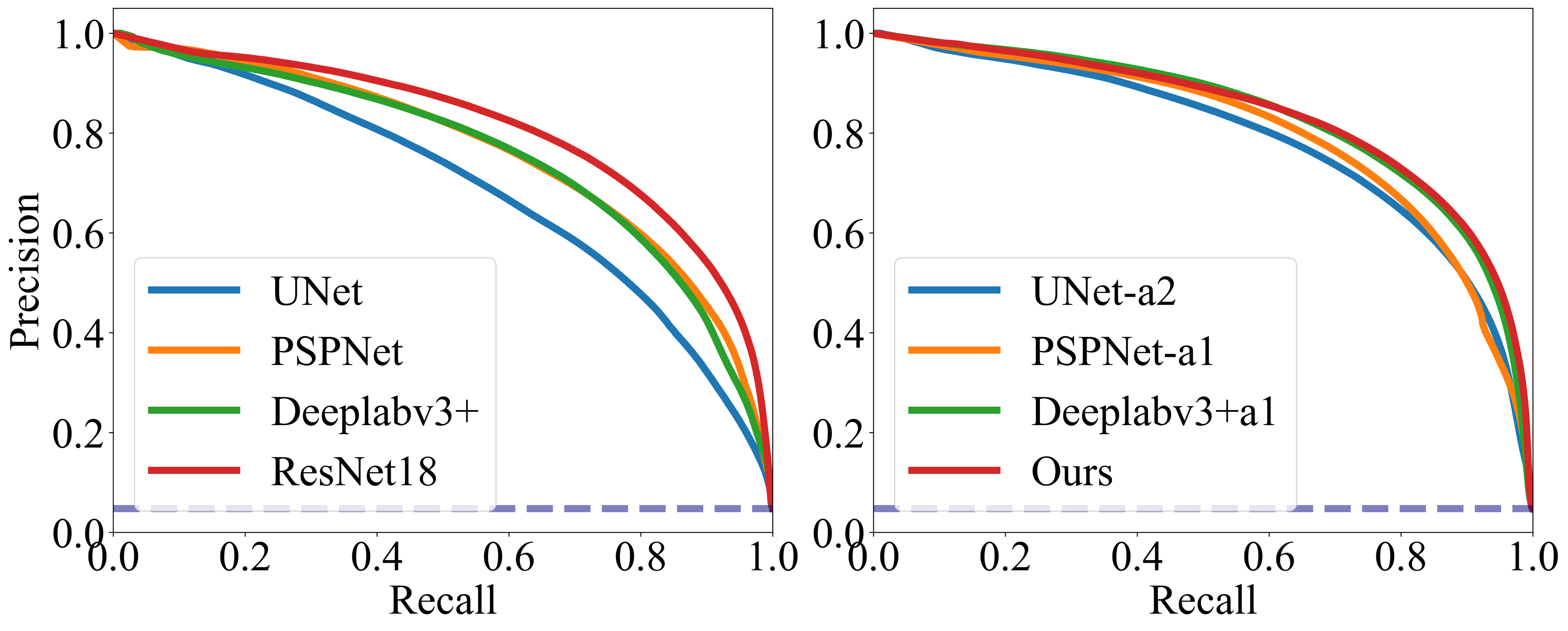}
    \caption{Precision-Recall curves comparison between models w/ and w/o attention gate and atrous blocks. The dashed navy-colored lines are baselines (0.043), which indicates an extreme class imbalance.}
    \label{fig:PR_curve}
    \vspace{-3mm}
\end{figure}

For ablation study on the attention gate and atrous blocks architecture, they achieve, on average across the four models, a 10.3\% and 10.58\% improvement in AUROC and AP over standard architecture, with MaxF and REC improving by 17.35\% and 30.10\%. Notably, the models have larger FPR due to the proposed architecture. Due to severe class imbalance, we employ precision-recall (PR) analysis rather than receiver operating characteristic (ROC) to evaluate model performance. Fig. \ref{fig:PR_curve} presents an intuitive comparison of the ablation study.

To evaluate the feasibility of our framework for real-time applications, we profiled its execution latency across all pipeline stages (Table \ref{tab:latency_profile}) using an NVIDIA RTX 5090 GPU on an Ubuntu 22.04 LTS system. The preprocessing pipeline, comprising multi-modal downsampling, CLAHE, and static homography alignment, requires an average of only $3.89\text{ ms}$ ($257.1\text{ Hz}$). Notably, this phase accounts for less than $11\%$ of the total end-to-end latency, ensuring that the vast majority of the computational budget is preserved for downstream neural network inference.

We also analyzed the computational trade-offs introduced by our architectural enhancements across all candidate backbones. As shown in Table \ref{tab:latency_profile}, these modules expectedly increase average inference latency; for example, the baseline ResNet18 architecture experiences a latency increase from $26.85\text{ ms}$ ($37.2\text{ FPS}$) to $34.73\text{ ms}$ ($28.8\text{ FPS}$) when integrated with the attention and atrous blocks. Nevertheless, the end-to-end framework maintains a throughput of $25.9\text{ FPS}$ ($38.62\text{ ms}$). This processing rate satisfies the standard real-time requirement ($15\text{--}30\text{ Hz}$) for autonomous driving applications.

\begin{table}[t]
	\caption{Latency and Throughput Profile of the Processing Pipeline and Models}
	\label{tab:latency_profile}
	\centering
	\setlength{\tabcolsep}{6pt} 
	\fontsize{8.8pt}{10.5pt}\selectfont
	\begin{tabularx}{\columnwidth}{Xcc}
		\toprule[1.0pt]
		\textbf{Pipeline Stage / Model Variant} & \textbf{Latency (ms)} & \textbf{Throughput} \\ \midrule
		\textit{Preprocessing Pipeline} & & \\
		Stage 1: Downsample (RGB) & 0.09 & - \\
		Stage 1: Downsample (Thermal) & 0.01 & - \\
		Stage 2: CLAHE (RGB) & 2.10 & - \\
		Stage 2: CLAHE (Thermal) & 0.95 & - \\
		Stage 3: Merge & 0.74 & - \\
		\textbf{Total Preprocessing} & \textbf{3.89} & \textbf{257.1 Hz} \\ \midrule
		\textit{Model Inference (Stage 4)} & & \\
		UNet & 23.43 & 42.7 FPS \\
		UNet + a2 & 29.78 & 33.6 FPS \\
		DeepLabv3+ & 26.33 & 38.0 FPS \\
		DeepLabv3+a1 & 33.15 & 30.2 FPS \\
		PSPNet & 25.26 & 39.6 FPS \\
		PSPNet+a1 & 30.54 & 32.8 FPS \\
		ResNet18 (Baseline) & 26.85 & 37.2 FPS \\
		\textbf{ResNet18+a2 (Proposed)} & \textbf{34.73} & \textbf{28.8 FPS} \\ \midrule
		\textbf{End-to-End System (Proposed)} & \textbf{38.62} & \textbf{25.9 FPS}{} \\ \bottomrule[1.0pt]
	\end{tabularx}
	\vspace{-3mm}
\end{table}

\subsection{Discussion}

Our results demonstrate that the fused RGB-I approach achieves the best overall performance, exceeding RGB-only and thermal-only. Also, integrating attention gates and atrous blocks yields consistent performance gains across all evaluated baseline models. Finally, we acknowledge certain limitations regarding our work: (1) while the metrics are promising, their sufficiency for a robust autonomous driving stack depends heavily on how perception errors affect downstream trajectory planning, especially given low-friction winter dynamics; (2) the current sample size is relatively modest, reflecting the inherent difficulty of gathering synchronized multi-modal data in extreme winter conditions; and (3) the ThermalTrack dataset captures specific winter conditions and may not represent the entire spectrum of snow types such as deep powder, ice crusts, or melting slush, each of which alters the visual and thermal signatures of wheel tracks. Consequently, our results should be interpreted as a validation of the architectural framework under these specific environmental constraints.

%% file: sections/conclusion.tex
This work presents a novel self-supervised pipeline for wheel track detection in snow-covered environments. Unlike existing drivable-area or traversability estimation methods - which often predict overly broad, unsafe regions by disregarding the critical benefits of following existing wheel tracks (e.g., compacted snow stability, reduced slippage, and validated path integrity) - our approach explicitly leverages these winter-specific navigational cues to enhance safety and reliability in featureless winter terrain. Experimental results demonstrate that thermal imaging reduces the false negative rate by 69.74\% compared to RGB-only baselines while maintaining a low false positive rate (0.014), providing critical discriminative features for track detection in snow-covered environments. Our RGB-thermal fusion pipeline achieves superior performance, validating the utility of multimodal sensing for winter navigation.

While this work addresses critical gaps in winter autonomy through low-cost and efficient RGB-thermal fusion, future directions include: (1) environmental scalability: investigating domain adaptation techniques for more diverse terrains (e.g., slush, ice patches, mud pits, flooded roads) and varied illumination or weather conditions (e.g., day/night settings, snowfall, fog, or headlight glare); (2) vehicle integration: combining perception with vehicle dynamics models to refine track quality estimation; (3) temporal modeling: leveraging recurrent modules or spatio-temporal transformers to exploit sequential frame dependencies and enhance tracking accuracy and continuity; and (4) digital twins: utilizing high-fidelity simulation environments to generate highly diverse winter scenarios and evaluate system performance across a broader spectrum of edge-case road dynamics.

%% file: bibliography/references.bib
@ARTICLE{9115644,
  author={Ruotsalainen, Laura and Renaudin, Valerie and Pei, Ling and Piras, Marco and Marais, Juliette and Cavalheri, Emerson and Kaasalainen, Sanna},
  journal={IEEE Intelligent Transportation Systems Magazine}, 
  title={Toward Autonomous Driving in Arctic Areas}, 
  year={2020},
  volume={12},
  number={3},
  pages={10-24},
  doi={10.1109/MITS.2020.2994014}}

@techreport{USDOT_AVAW2021,
  author      = "{U.S. Department of Transportation, Federal Highway Administration}",
  title       = "{Automated Vehicles and Adverse Weather (AVAW) Phase 3-Final Report}",
  institution = {},
  year        = {2021},
  month       = {},
  address     = {},
  url         = {https://ops.fhwa.dot.gov/publications/fhwahop21047/fhwahop21047.pdf}
}

@ARTICLE{8931615,
  author={Morad, Steven D. and Nash, Jeremy and Higa, Shoya and Smith, Russell and Parness, Aaron and Barnard, Kobus},
  journal={IEEE Robotics and Automation Letters}, 
  title={Improving Visual Feature Extraction in Glacial Environments}, 
  year={2020},
  volume={5},
  number={2},
  pages={385-390},
  doi={10.1109/LRA.2019.2959490}}

@article{Baril_2022,
   title={Kilometer-Scale Autonomous Navigation in Subarctic Forests: Challenges and Lessons Learned},
   volume={2},
   ISSN={2771-3989},
   url={http://dx.doi.org/10.55417/fr.2022050},
   DOI={10.55417/fr.2022050},
   journal={Field Robotics},
   publisher={Institute of Electrical and Electronics Engineers (IEEE)},
   author={Baril, Dominic and Deschênes, Simon-Pierre and Gamache, Olivier and Vaidis, Maxime and LaRocque, Damien and Laconte, Johann and Kubelka, Vladimír and Giguère, Philippe and Pomerleau, François},
   year={2022},
   month=jul, pages={1628–1660} }

@INPROCEEDINGS{8594031,
  author={Hirose, Noriaki and Sadeghian, Amir and Vázquez, Marynel and Goebel, Patrick and Savarese, Silvio},
  booktitle={2018 IEEE/RSJ International Conference on Intelligent Robots and Systems (IROS)}, 
  title={GONet: A Semi-Supervised Deep Learning Approach For Traversability Estimation}, 
  year={2018},
  volume={},
  number={},
  pages={3044-3051},
  doi={10.1109/IROS.2018.8594031}}

@INPROCEEDINGS{9561394,
  author={Lee, Hyunsuk and Chung, Woojin},
  booktitle={2021 IEEE International Conference on Robotics and Automation (ICRA)}, 
  title={A Self-Training Approach-Based Traversability Analysis for Mobile Robots in Urban Environments}, 
  year={2021},
  volume={},
  number={},
  pages={3389-3394},
  doi={10.1109/ICRA48506.2021.9561394}}

@ARTICLE{9839522,
  author={Gasparino, Mateus V. and Sivakumar, Arun N. and Liu, Yixiao and Velasquez, Andres E. B. and Higuti, Vitor A. H. and Rogers, John and Tran, Huy and Chowdhary, Girish},
  journal={IEEE Robotics and Automation Letters}, 
  title={WayFAST: Navigation With Predictive Traversability in the Field}, 
  year={2022},
  volume={7},
  number={4},
  pages={10651-10658},
  doi={10.1109/LRA.2022.3193464}}

@article{kurup2023winter,
  title={Winter adverse driving dataset for autonomy in inclement winter weather},
  author={Kurup, Akhil M and Bos, Jeremy P},
  journal={Optical Engineering},
  volume={62},
  number={3},
  pages={031207--031207},
  year={2023},
  publisher={Society of Photo-Optical Instrumentation Engineers}
}

@book{ordonez2010rut,
  title={Rut detection and following for autonomous ground vehicles},
  author={Ordonez, Camilo},
  year={2010},
  publisher={The Florida State University}
}

@article{ordonez2011laser,
  title={Laser-based rut detection and following system for autonomous ground vehicles},
  author={Ordonez, Camilo and Chuy Jr, Oscar Y and Collins Jr, Emmanuel G and Liu, Xiuwen},
  journal={Journal of Field Robotics},
  volume={28},
  number={2},
  pages={158--179},
  year={2011},
  publisher={Wiley Online Library}
}

@inproceedings{schmid2022self,
  title={Self-supervised traversability prediction by learning to reconstruct safe terrain},
  author={Schmid, Robin and Atha, Deegan and Sch{\"o}ller, Frederik and Dey, Sharmita and Fakoorian, Seyed and Otsu, Kyohei and Ridge, Barry and Bjelonic, Marko and Wellhausen, Lorenz and Hutter, Marco and others},
  booktitle={2022 IEEE/RSJ International Conference on Intelligent Robots and Systems (IROS)},
  pages={12419--12425},
  year={2022},
  organization={IEEE}
}

@article{alamikkotervo2024tadap,
  title={TADAP: Trajectory-Aided Drivable area Auto-labeling with Pretrained self-supervised features in winter driving conditions},
  author={Alamikkotervo, Eerik and Ojala, Risto and Sepp{\"a}nen, Alvari and Tammi, Kari},
  journal={IEEE Transactions on Intelligent Vehicles},
  year={2024},
  publisher={IEEE}
}

@ARTICLE{10146445,
  author={Seo, Junwon and Sim, Sungdae and Shim, Inwook},
  journal={IEEE Robotics and Automation Letters}, 
  title={Learning Off-Road Terrain Traversability With Self-Supervisions Only}, 
  year={2023},
  volume={8},
  number={8},
  pages={4617-4624},
  doi={10.1109/LRA.2023.3284356}}

@inproceedings{fuentes2021quantification,
  title={Quantification of Rut Detection and Height Mapping in Winter Terrains for Off-Road Mobility},
  author={Fuentes, Anthony J and Vecherin, Sergey N and Bodie, Mark O and Parker, Michael W},
  booktitle={Regional Conference on Permafrost 2021 and the 19th International Conference on Cold Regions Engineering},
  pages={172--183},
  year={2021},
  organization={American Society of Civil Engineers Reston, VA}
}

@article{zhang2024identification,
  title={Identification of wheel track in the wheat field},
  author={Zhang, Wanhong},
  journal={Scientific Reports},
  volume={14},
  number={1},
  pages={900},
  year={2024},
  publisher={Nature Publishing Group UK London}
}

@article{seok2024lidar,
  title={LiDAR-based Road Height Profile Estimation and Bump Detection for Preview Suspension},
  author={Seok, Jiwon and Jo, Jaeyoung and Kim, Youngjae and Kim, Hyungjin and Jung, Inyong and Jeong, Mugwan and Kim, Namgyun and Jo, Kichun},
  journal={IEEE Transactions on Intelligent Vehicles},
  year={2024},
  publisher={IEEE}
}

@article{goberville2022tire,
  title={Tire track identification: A method for drivable region detection in conditions of snow-occluded lane lines},
  author={Goberville, Nicholas A and Kadav, Parth and Asher, Zachary D},
  journal={SAE International Journal of Advances and Current Practices in Mobility},
  volume={4},
  number={2022-01-0083},
  pages={1590--1597},
  year={2022}
}

@INPROCEEDINGS{7487782,
  author={Sock, Juil and Kim, Jun and Min, Jihong and Kwak, Kiho},
  booktitle={2016 IEEE International Conference on Robotics and Automation (ICRA)}, 
  title={Probabilistic traversability map generation using 3D-LIDAR and camera}, 
  year={2016},
  volume={},
  number={},
  pages={5631-5637},
  doi={10.1109/ICRA.2016.7487782}}

@INPROCEEDINGS{7759199,
  author={Wermelinger, Martin and Fankhauser, Péter and Diethelm, Remo and Krüsi, Philipp and Siegwart, Roland and Hutter, Marco},
  booktitle={2016 IEEE/RSJ International Conference on Intelligent Robots and Systems (IROS)}, 
  title={Navigation planning for legged robots in challenging terrain}, 
  year={2016},
  volume={},
  number={},
  pages={1184-1189},
  doi={10.1109/IROS.2016.7759199}}

@inproceedings{shaban2022semantic,
  title={Semantic terrain classification for off-road autonomous driving},
  author={Shaban, Amirreza and Meng, Xiangyun and Lee, JoonHo and Boots, Byron and Fox, Dieter},
  booktitle={Conference on Robot Learning},
  pages={619--629},
  year={2022},
  organization={PMLR}
}

@InProceedings{Zhao_2017_CVPR,
author = {Zhao, Hengshuang and Shi, Jianping and Qi, Xiaojuan and Wang, Xiaogang and Jia, Jiaya},
title = {Pyramid Scene Parsing Network},
booktitle = {Proceedings of the IEEE Conference on Computer Vision and Pattern Recognition (CVPR)},
month = {July},
year = {2017}
}

@inproceedings{barnes2017find,
  title={Find your own way: Weakly-supervised segmentation of path proposals for urban autonomy},
  author={Barnes, Dan and Maddern, Will and Posner, Ingmar},
  booktitle={2017 IEEE International Conference on Robotics and Automation (ICRA)},
  pages={203--210},
  year={2017},
  organization={IEEE}
}

@ARTICLE{8963641,
  author={Wellhausen, Lorenz and Ranftl, René and Hutter, Marco},
  journal={IEEE Robotics and Automation Letters}, 
  title={Safe Robot Navigation Via Multi-Modal Anomaly Detection}, 
  year={2020},
  volume={5},
  number={2},
  pages={1326-1333},
  doi={10.1109/LRA.2020.2967706}}

@inproceedings{xaud2019thermal,
  title={Thermal image based navigation system for skid-steering mobile robots in sugarcane crops},
  author={Xaud, Marco FS and Leite, Antonio C and From, P{\aa}l J},
  booktitle={2019 International Conference on Robotics and Automation (ICRA)},
  pages={1808--1814},
  year={2019},
  organization={IEEE}
}

@inproceedings{aditya2024thermal,
  title={Thermal voyager: A comparative study of rgb and thermal cameras for night-time autonomous navigation},
  author={Aditya, NG and Dhruval, PB and Shalabi, Jehan and Jape, Shubhankar and Wang, Xueji and Jacob, Zubin},
  booktitle={2024 IEEE International Conference on Robotics and Automation (ICRA)},
  pages={14116--14122},
  year={2024},
  organization={IEEE}
}

@Article{s24237802,
AUTHOR = {Kim, Donghyun and Jeong, Yonghwan},
TITLE = {Development of Robust Lane-Keeping Algorithm Using Snow Tire Track Recognition in Snowfall Situations},
JOURNAL = {Sensors},
VOLUME = {24},
YEAR = {2024},
NUMBER = {23},
ARTICLE-NUMBER = {7802},
URL = {https://www.mdpi.com/1424-8220/24/23/7802},
PubMedID = {39686338},
ISSN = {1424-8220},
DOI = {10.3390/s24237802}
}

@inproceedings{he2016deep,
  title={Deep residual learning for image recognition},
  author={He, Kaiming and Zhang, Xiangyu and Ren, Shaoqing and Sun, Jian},
  booktitle={Proceedings of the IEEE conference on computer vision and pattern recognition},
  pages={770--778},
  year={2016}
}

@inproceedings{zhao2017pyramid,
  title={Pyramid scene parsing network},
  author={Zhao, Hengshuang and Shi, Jianping and Qi, Xiaojuan and Wang, Xiaogang and Jia, Jiaya},
  booktitle={Proceedings of the IEEE conference on computer vision and pattern recognition},
  pages={2881--2890},
  year={2017}
}

@inproceedings{ronneberger2015u,
  title     = {U-Net: Convolutional Networks for Biomedical Image Segmentation},
  author    = {Ronneberger, O. and Fischer, P. and Brox, T.},
  booktitle = {Medical image computing and computer-assisted intervention--MICCAI 2015},
  year      = {2015},
  pages     = {234--241},
  publisher = {Springer}
}

@inproceedings{chen2018encoder,
  title={Encoder-decoder with atrous separable convolution for semantic image segmentation},
  author={Chen, Liang-Chieh and Zhu, Yukun and Papandreou, George and Schroff, Florian and Adam, Hartwig},
  booktitle={Proceedings of the European conference on computer vision (ECCV)},
  pages={801--818},
  year={2018}
}

@inproceedings{lin2017focal,
  title={Focal loss for dense object detection},
  author={Lin, Tsung-Yi and Goyal, Priya and Girshick, Ross and He, Kaiming and Doll{\'a}r, Piotr},
  booktitle={Proceedings of the IEEE international conference on computer vision},
  pages={2980--2988},
  year={2017}
}

@inproceedings{kervadec2019boundary,
  title={Boundary loss for highly unbalanced segmentation},
  author={Kervadec, Hoel and Bouchtiba, Jihene and Desrosiers, Christian and Granger, Eric and Dolz, Jose and Ayed, Ismail Ben},
  booktitle={International conference on medical imaging with deep learning},
  pages={285--296},
  year={2019},
  organization={PMLR}
}

@article{oktay2018attention,
  title={Attention u-net: Learning where to look for the pancreas},
  author={Oktay, Ozan and Schlemper, Jo and Folgoc, Loic Le and Lee, Matthew and Heinrich, Mattias and Misawa, Kazunari and Mori, Kensaku and McDonagh, Steven and Hammerla, Nils Y and Kainz, Bernhard and others},
  journal={arXiv preprint arXiv:1804.03999},
  year={2018}
}

@article{vachmanus2021multi,
  title={Multi-modal sensor fusion-based semantic segmentation for snow driving scenarios},
  author={Vachmanus, Sirawich and Ravankar, Ankit A and Emaru, Takanori and Kobayashi, Yukinori},
  journal={IEEE sensors journal},
  volume={21},
  number={15},
  pages={16839--16851},
  year={2021},
  publisher={IEEE}
}

@INPROCEEDINGS{9294671,
  author={Humblot-Renaux, Galadrielle and Li, Vivian and Pinto, Daniela and Marchegiani, Letizia},
  booktitle={2020 IEEE 23rd International Conference on Intelligent Transportation Systems (ITSC)}, 
  title={Thermal Imaging on Smart Vehicles for Person and Road Detection: Can a Lazy Approach Work?}, 
  year={2020},
  volume={},
  number={},
  pages={1-6},
  doi={10.1109/ITSC45102.2020.9294671}}
